\documentclass[sigconf]{acmart}
\usepackage{amsmath}
\usepackage[toc,page]{appendix} 

\usepackage{algorithm}
\usepackage{bm}
\usepackage[noend]{algpseudocode}

\newcommand{\WRP}{\par\qquad\(\hookrightarrow\)\enspace}

\AtBeginDocument{%
  \providecommand\BibTeX{{%
    \normalfont B\kern-0.5em{\scshape i\kern-0.25em b}\kern-0.8em\TeX}}}

\acmConference[]{tbd}{tbd}{tbd}

\begin{document}

\title{Keyword-based Topic Modeling and Keyword Selection}


%
\author{Xingyu Wang, Lida Zhang, Diego Klabjan} 
       \affiliation{Northwestern University} 
       \affiliation{Evanston, Illinois}
       \email{xingyuwang2017,lidazhang2018@u.northwestern.edu,d-klabjan@northwestern.edu} 


\renewcommand{\shortauthors}{WZK}

\begin{abstract}
  Certain type of documents such as tweets are collected by specifying a set of keywords. As topics of interest change with time it is beneficial to adjust keywords dynamically. The challenge is that these need to be specified ahead of knowing the forthcoming documents and the underlying topics. The future topics should mimic past topics of interest yet there should be some novelty in them. We develop a keyword-based topic model that dynamically selects a subset of keywords to be used to collect future documents. The generative process first selects keywords and then the underlying documents based on the specified keywords. The model is trained by using a variational lower bound and stochastic gradient optimization. The inference consists of finding a subset of keywords where given a subset the model predicts the underlying topic-word matrix for the unknown forthcoming documents. We compare the keyword topic model against a benchmark model using viral predictions of tweets combined with a topic model. The keyword-based topic model outperforms this sophisticated baseline model by 67\%.
\end{abstract}




\maketitle

\section{Introduction}
Keywords are an important means to capture most significant information in a text document. They are used in information retrieval, text summarization, social media streaming, etc.  
Twitter offers an Application Programming Interface (API) for streaming of tweets. The API supports many options to filter tweets based on several streaming endpoints, such as language, location, users, and keywords. Filtering by keywords is a very popular and efficient way to capture tweets based on target interest and domain. With the keyword filter, a tweet is captured if it includes any keyword from the keyword list. Many companies use the names of the company or products to monitor customer feedback, provided customer support, conduct focus groups, and engage in similar activities. However, this approach of collecting feedback has limitations. Since the keywords are selected in advance and then kept static for an extended period of time, a useful tweet containing valuable user feedback but not including any of the selected keywords cannot be captured. For example, consider a bank using its name as the keyword to monitor online feedback about their new credit cards. If a Twitter user posts an image of a credit card as a tweet and comments, ``I dislike the rewards of this credit card,'' then this tweet would not be captured, even though it contains very useful information. Such cases frequently occur in replies to posted tweets or comments of retweets because the company or product names might already be mentioned in the original tweet and are not captured by the keyword. If the set of keywords is expanded by including “credit card,” then too many tweets would be captured. There is even a bigger fundamental flaw. In the near future, there might be a relevant event or activity coming up leading to a new topic that would not be captured by prior keywords. This calls for having a prediction model for the forthcoming topics and dynamically adjusting the keywords to anticipate emerging topics. Indeed, this research results from a collaboration with a large global company having challenges updating keywords on a regular basis. 

As input the problem consists of a set of documents with a temporal dimension in a corpus and a set of keywords used to collect each document. Each document includes at least one keyword. The documents collected in a time window form a set of topics. The problem is to select a set of new keywords that are going to be used to collect documents in the immediate future. The future topics which have to be predicted from the keywords implied by the yet-to-be-collected documents should not deviate too much from the recent topics, but, on the other hand, should capture possible emerging topics. 

Word frequency is a straight forward approach to extract new keywords. TFIDF is an early word extraction algorithm comparing word frequency in a document and in the corpus to determine the importance of a word. There are extensions to this notion based on TFIDF that apply extra features, such as the length of the words, however, tweets are short (maximum 280 character posts) and thus it is hard to find meaningful words that frequently appear in certain documents but have low frequency in others. With TFIDF, it is also easy to pick background words that seem useful but are too broad to be keywords, such as ``today,” ``money,” etc. With such broad words as keywords, it is easy to filter short tweets that include these words but are off-topic. Thus we need a different algorithm and model to select meaningful keywords applicable to short text media like Twitter.
In addition, TFIDF and its derived algorithms select keywords based on historical data, so the selected keywords can only represent past information. However, new keywords need to be used in the future instead of being based on the past. To capture useful information, we need to select keywords that should encapsulate upcoming topics and trends. Therefore, a generative model is necessary for our problem in addition to an inference part that is going to predict upcoming topics based on sets of keywords. Unfortunately, TFIDF does not provide such a generative model.

To select meaningful keywords that can be used to capture future relevant topics, we propose a generative model with a topic model as an information filter where priors depend on candidate sets of keywords. There are some keyword extraction models based on topics, such as Topical PageRank \cite{liu2010automatic} and context-sensitive Topical PageRank \cite{zhao2011topical}. These models calculate word ranking scores using the graph-based Topical PageRank and extract new keywords from the word ranking scores. They build word graphs and use topics to measure the importance of the words. In such models topics are fixed, and the keywords within the chosen topics are subsequently selected. Since the topics are fixed, these models do not meet our requirement of anticipating and predicting future active keywords together with resulting topics. In our proposed model, both topics and keywords have equal importance and are jointly modeled with the generative model. The generative model captures past interactions between the keywords and topics and thus in inference given a set of keywords it predicts the resulting topics. A challenge in inference is to balance the fact that future topics should be similar to the current topics, yet at the same time they should capture emerging topics. 

Different from Topical PageRank and context-sensitive Topical PageRank, which are limited to pre-selected and fixed topics, we model topics by a topic model (Latent Dirichlet Allocation (LDA), \cite{blei2003latent}). The topics are modeled in time and depending on the underlying keywords, and thus can reflect events and trends, such as a new product release. We assume that the set of all documents is grouped into subsets (e.g. all documents collected in a week) where each subset of documents is generated based on its own set of keywords. The keywords pertaining to the different subsets can overlap. We model the interaction between keywords and topics through priors at the document-topic and topic-word levels. The keyword selection process has its own simple generative process whose output forms the input to neural networks that in turn output the priors to LDA. 
The model has three prior parameters $\gamma$, $\lambda$, and $\psi$ which are all trainable. In each updating period (e.g. a week) corresponding to a subset of documents and the underlying keywords, the selected keywords are first sampled from prior parameter $\gamma$. The generated keywords work together with prior parameter $\psi$ (weights in a neural network) to generate the document-topic distribution $\eta$ through a neural network, and with the other prior parameter $\lambda$ (also weights in a neural network) through another neural network to generate the topic-word distribution $\phi$ indicating the word distribution for each topic. Regarding word generation in document $d$, first latent topic variable $z_d^{kw}$ is selected based on the generated set of keywords and next the keyword belonging to this document is sampled from the topic-word distribution $\psi$ (which also depends on the generated keywords through a neural network). The latter sampling within the topic selected is based on a discrete probability conditioned on topic $z_d^{kw}$ and the topic-word distribution. For each remaining word in the document, latent topic $z_{w,d}$ is sampled from the topic-word distribution $\psi$, and a word is chosen from discrete probability conditioned on topic $z_{w,d}$ (this step follows standard LDA). In this way, every word and keyword for each document are sampled. The process guarantees that each generated document has at least one word from the generated set of keywords. 

The model includes the standard LDA parameters which are augmented by the parameters pertaining to the generative process behind keyword generation and it also includes neural network parameters. The parameters are trained by a combination of expectation maximization (EM), variational inference, and stochastic gradient optimization. Since the generative process behind keywords and a part of the incumbent LDA adjusted model are using discrete distributions, we also resort to the Gumbel softmax trick in training. In inference, we need to decide which keywords to use for immediate future streaming based on the generative model. 

The proposed model is evaluated on three sets of tweets with the benchmark algorithm being a sophisticated algorithm based on NLP techniques (also designed in this work). In terms of accuracy appropriately defined the improvement is 74\%, 61\% over the benchmark algorithm if 2 or 3 keywords are recommended, respectively. If a random subset of keywords is selected among all candidates words, then the improvement is 3-fold. 

In this paper, we propose a novel generative probabilistic model for keyword selection, Keyword Latent Dirichlet Allocation (KLDA). The main contributions of our work as as follows:
\begin{enumerate}
\item A topic based generative model is proposed for the future document keyword selection.
\item Neural networks are introduced in producing priors.
\item An inference model that recommends keywords to use in the future trading off the facts that future topics should be similar to the current topics, yet they should include novelty in order not to miss emerging topics. 
\end{enumerate}

In Section 2 we review the literature while in Section 3 the problem is formally stated. In Section 4, we describe the details of KLDA, including the generative process and inference. In Section 5 we introduce the datasets and discuss all of the experimental results. 

\section{Literature Review}
\subsection{Keyword Extraction}
Keywords are an efficient and common way of filtering streaming textual data (in particular on Twitter). Several prior works deal with keyword extraction. Turney \cite{turney2000learning} proposes a supervised learning approach for this problem, classifying phrases as positive and negative in each document. Many later algorithms have been developed based on the Turney's approach, with additional features and data sources. Hulth \cite{hulth2003improved} introduces part-of-speech (POS) tags to represent the data, Yih et al. \cite{yih2006finding} analyze information from many resources, including meta-tags, URL, and query frequency, and Liu et al. \cite{liu2008automatic} utilize a rich set of features such as lexical. However, deciding the importance of a word based on its frequency does not capture the resulting topics. In addition, supervised learning methods require accurate learning labels which are usually not available. The goal is to periodically set new keywords for future upcoming documents, so it is hard to get labels for a massive amount of documents. Besides, we do not have a clear target for what the keywords should be. Therefore, unsupervised learning is necessary in our keyword selection problem. 

Graph-based keyword ranking is a popular and effective strategy to extract keywords. PageRank \cite{page1999pagerank} is proposed by Page and Brin to estimate the importance of Web pages. Relying on PageRank, Mihalcea and Tarau \cite{mihalcea2004textrank} propose a keyword extraction method based on their TextPage model. TextPage is widely used in later keyword extraction algorithms \cite{wan2007towards, liu2009unsupervised}. However, these graph-based models do not consider the semantics of keywords. Thus, Liu et al. and Zhao et al. \cite{liu2010automatic, zhao2011topical} propose topic based PageRank focusing on keyword extraction in pre-selected and fixed topics. However, we aim to select keywords without knowing the topics a priori and thus we rely on the concept of predicting future topics. 


\subsection{Topic Models}
Topic models discover semantic relationships in documents by maintaining latent topics of documents. Latent Semantic Indexing  \cite{papadimitriou2000latent} is an early topic model that maps words and documents to a representation in a semantic space using Singular Value Decomposition assuming that the words with the same semantics have similar meanings. Hofmann \cite{hofmann2017probabilistic} propose a generative model Probabilistic Latent Semantic Indexing that contains a statistical foundation to build document indexing and information retrieval. It obtains the probabilities of different topics by generating every word from a topic and thus each document may contain multiple topics. Based on this, Blei et al. \cite{blei2003latent} introduce LDA to overcome overfitting of Latent Semantic Indexing models. A survey of topic models can be found in \cite{daud2010knowledge}.

LDA is a generative model for detecting topics. It represents each topic with the word distribution and models every document as a mixture of multiple latent topics. Many models consider other factors, and extend LDA in different dimensions.  
Blei et al. \cite{blei2006dynamic} consider topic evolution in sequential documents and extends the topic model to the dynamic version, and Hoffman et al. \cite{hoffman2010online} extrapolate LDA from local analysis to online learning in streams.
In image problems, Wang et al. \cite{wang2008spatial} introduce image information into LDA priors to encode the spatial structure among visual cues, Li et al. \cite{fei2005bayesian} extend LDA to natural scene categories, and Barnard et al. \cite{barnard2003matching} also apply LDA parameters on images for matching words and pictures. Lin et al. and Mass et al.  \cite{lin2009joint, maas2011learning} generalize LDA in sentiment analyses, and Zhang et al., Steyvers et al., and Rosen-Zvi et al. \cite{zhang2015prior, steyvers2004probabilistic, rosen2004author} add user information into LDA priors. 
Similar to these works, we introduce keywords into the LDA model by means of a flexible prior. The approaches are fundamentally different because of the dependence of topics on the underlying keywords.

\begin{figure*}
  \centering
  {\includegraphics[width=.9\textwidth]{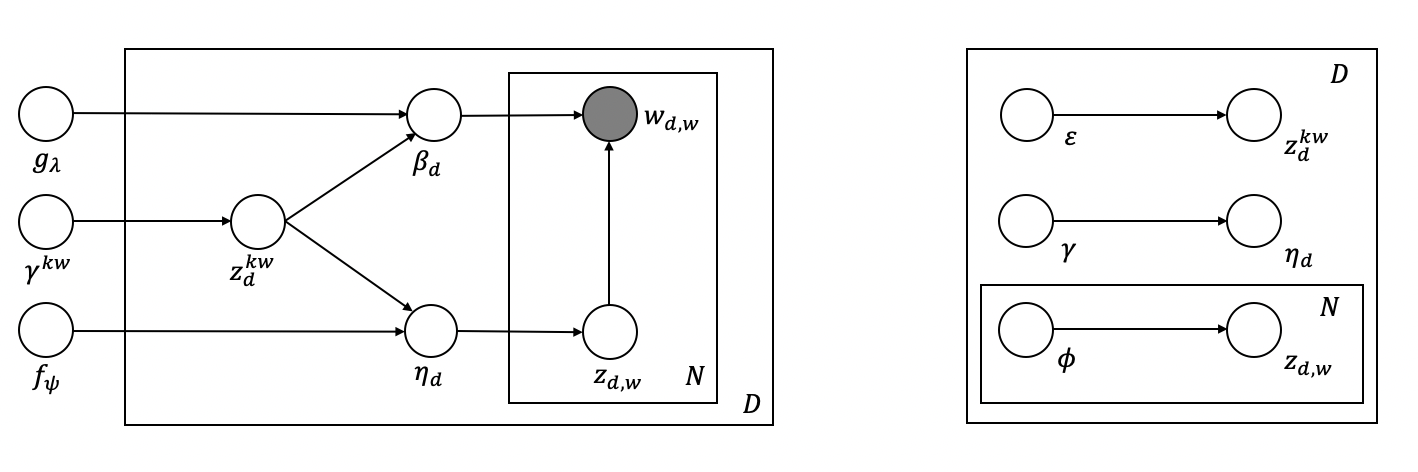}\label{fig:exph}}
  \caption{(Left) Plate representation of KLDA. (Right) Graphical model Representation for the variational distribution approximating posterior of KLDA. Subscript $d$ is used for variables that are realized at document level, while $w$ for variables generated for each observed word in the corpus.}
  \medskip
  \small
  \label{fig_graph_model}
\end{figure*}

\section{Model and Algorithm}

We first present notation and the setting. We assume that, from a finite vocabulary set $\mathcal{V} = \{W_1,W_2,\cdots,W_{V}\}$ (where $V$ represents the cardinality of the vocabolary), we are given a subset $\mathcal{Q} \subset \mathcal{V}$ called the keyword set, which is the focus of this work. We use $Q$ to denote the cardinality of $\mathcal{Q}$. A document $d$ is a set of words $\{W_{d,1},W_{d,2},\cdots,W_{d,{n_d}}\}$ where $n_d \in \mathbb{Z}$ is the size of the set, $W_{d,i}\in \mathcal{V}$ for all $i = 1,2,\cdots,n_d$, and there exists at least one $kw(d)\in \{1,2,\cdots,n_d\}$ such that $W_{d,kw(d)} \in \mathcal{Q}$. Note that (1) we require each document to contain at least one keyword; (2) there can be more than one keyword in a document; (3) we adopt the bag-of-words assumption throughout, thus the exact order of the words in the document is of no importance. A corpus $\mathcal{W}$ is a sequence of documents $(d_1,d_2,\cdots,d_{D})$ where we assume that the index reflects the temporal dimension, i.e., $d_1$ is the oldest document and $d_D$ the most recent document. We denote by $K$ the number of topics which is a fixed hyperparameter.

As input (training data) we are given $\mathcal{W}$ together with $\mathcal{Q}$. The main goal of keyword selection is to find a subset $\mathcal{Q}_{next}\subseteq \mathcal{Q}$ such that yet-to-be-collected documents $d_{D+1},d_{D+2},\cdots$ based on keywords $\mathcal{Q}_{next}$ have similar topics as the topics of the most recent documents in the corpus. ``Similarity'' is defined later in Section \ref{secInference}; it needs to capture the desire to have identical topics yet to also include some emerging relevant topics.   

As an example with tweets, let us assume that at the beginning of each week we specify a set of keywords. During a week we collect all tweets with respect to these keywords. Training data consists of all past tweets collected together with the keywords used. The problem is to select a set of keywords for the next week tweets. In this example, $\mathcal{W}$ corresponds to all historical tweets and $\mathcal{Q}$ can be the set of all keywords used in prior weeks possibly augmented by additional potential keywords. The keywords for the next week are denoted by $\mathcal{Q}_{next}$. The intent is that the topics of the tweets in the next week would be similar to the topics in most recent weeks but they should also capture some emerging relevant topics. 

We approach this problem from the perspective of topic modeling. Each week of tweets yields a set of topics where the document-topic and topic-word probabilities depend on keywords used. We model this dependency through priors of the underlying Dirichlet distributions. This yields a relationship between a potential set of keywords and topics for the next week. 

We first introduce the generative model and then the solution methodology based on variational inference. 

\subsection{Generative Model}

We propose the KLDA model (illustrated in Figure \ref{fig_graph_model} left) to explain the generative process of a document (labeled with subscript $d$ and containing $N$ words) with keywords, the steps of which can be summarized as follows.
\begin{enumerate}
\item Keyword generation: Sample $z^{kw}_d \sim p_{kw}(\ \cdot\ | \theta)$
\item Document keywords: $d = \{w:(z^{kw}_d)_w = 1\}$
\item Topic weights: Sample $\eta_d \sim \text{Dir}\big(f_{\psi}(z^{kw}_d)\big)$
\item Topic-word matrix parameters: $\beta_d = g_{\lambda}(z^{kw}_d)$
\item Let $n^{kw}_d=|\{w:(z^{kw}_d)_w = 1\}|$
\item For each of the remaining $N-n^{kw}_d$ words $W_{d,w}$
\begin{align*}
    & (a) \text{ Topic label: Sample }z_{d,w} \sim \text{Multin}(\eta_d) \\
    & (b) \text{ Generate word: Sample }W_{d,w} \sim p_{w} (\  \cdot\  | z_{d,w};\beta_d ) \\
    & (c) \text{ Add word to document: } d=d\cup \{W_{d,w}\} 
\end{align*}
\end{enumerate}


Notation ``Dir,''``Multin'' represent the Dirichlet and multinomial distributions, respectively. The trainable parameters are $\theta, \psi,\lambda$. In the exposition for simplicity we assume that $N$ is fixed, but everything can be easily extended to the case when $N$ is distributed based on the Poisson distribution.

Step 1 determines which of the words in the candidate keyword set $\mathcal{Q}$ are actually used as keywords for the current document. Binary $z^{kw}_d \in \{0,1\}^Q\setminus \{0\}$ represents the inclusion of the corresponding keyword in $d$ (1 if the keyword is in $d$ and 0 otherwise). These keywords become part of the document (step 2). Note that $z^{kw}_d\ne 0$ which means that each document must contain at least one keyword. We assume an energy-based distribution for keywords, where the unnormalized probability is $\tilde{p}_{kw}(z | \theta) = \exp(z^T\theta - c (\mathbf{1}^Tz - 1))$ with $c$ being a hyperparameter. The partition function $Z(\theta) = \sum_{\tilde{z}}\tilde{p}_{kw}(\tilde{z} | \theta)$ is a summation over all  possible combinations of $\tilde{z} \in \{0,1\}^Q \setminus \{0\}$. As the result, the normalized probability is $p_{kw}(z | \theta) = \tilde{p}_{kw}(z | \theta)/Z(\theta)$. In the energy function $\tilde{p}_{kw}(z | \theta)$ term $z^T\theta$ encourages the model to generate keywords that are aligned with $\theta$, which represents the underlying probability of each keyword while term $c (\mathbf{1}^Tz - 1)$ penalizes $z$ if too many keywords are included, which provides a regularization and prohibits the generation of a large number of keywords for a single document.

Next in step 3 we generate a probability vector $\eta_d$ of topics based on the Dirichlet distribution. Its prior 
$f_{\psi}$ is a model (parameterized by $\psi$) that takes as input a binary vector of dimension $Q$, and outputs a vector with $Q$ positive numbers that can be used as a prior for a Dirichlet distribution. In our model the prior depends on the topics generated $z^{kw}_d$.

In the next step 4 we generate a set of parameters $\beta_d$ that govern the topic-word distribution of the selected topic $z_{d,w}$. These parameters are specified by function $g_{\lambda}$ which is a model (parameterized by $\lambda$) that takes as input a binary vector of dimension $Q$ and outputs a $K\times V$ matrix with each row corresponding to a distribution that is used as a topic-word matrix in the LDA model.

In step 5, $n^{kw}_d$ is the number of keywords selected. Step 6 generates all of the remaining keywords in $d$.  In step 6a a topic is then selected based on the multinomial distribution with probabilities $\eta_d$. A word is finally sampled in step 6b based on a probability depending on the generated topic and topic-word priors, and added to the document in the last step. 
Distribution $p_w(\ \cdot\ |z;\beta)$ is a multinomial distribution conditioned on the one-hot vector $z$, meaning that for index $k$ with $z_k = 1$, the distribution is a multinomial distribution parametrized by the $k-$th row of $\beta$.

The generative process implies the joint probability distribution 
\begin{align*}
p(\mathcal{W},\mathbf{z}^{kw},\mathbf{\eta},\mathbf{z}& |\theta,\psi,\lambda) =  \prod_{d\in \mathcal{W}} p_{kw}(z^{kw}_d | \theta )\cdot p( \eta_d | z^{kw}_d, \psi ) \cdot
\\ & \prod_{w=1}^{n_d}  p(z_{d,w}|\eta_d) \cdot p_w(W_{d,w}|z_{d,w},\beta_d(\lambda)).
\end{align*}

One peculiarity of this generative process is that on the surface the topic-word matrix seem to depend on document $d$. However each document generated on the same set of keywords $z^{kw}$ is subject to the same topic-word matrix. This is consistent with our goal that a set of keywords implies a topic-word matrix. On the other hand, it is consistent with the standard LDA where there is a single set of keywords and thus just one such matrix not depending on documents. 

We describe the specific choices of $f,g$ and the procedures for the updates of their parameters in the next section.

\subsection{Variational Inference with EM Algorithm}

Due to the intractability of the proposed probabilistic model, we resort to variational inference combined with the EM algorithm for model training. The variational distribution for latent variables in KLDA is shown in Figure \ref{fig_graph_model} right and it reads
\begin{align*}
q(\mathbf{z}^{kw},\mathbf{\eta},\mathbf{z} |\epsilon, \gamma, \phi)
= & \prod_{d\in \mathcal{W}}
    q(z^{kw}_d| \epsilon_d) \cdot 
     q(\eta_d|\gamma_d ) \cdot \prod_{w}  q(z_{d,w}| \phi_{d,w}).
\end{align*}
In the study we assume that $q(z^{kw}_d| \epsilon_d)$ is the Bernoulli distribution with success rate $\epsilon_d$ which agrees with the assumption that multiple keywords can be presented in a document. Distribution $q(z_{d,w}| \phi_{d,w})$ of a given document $d$ is multinomial with parameters $\phi_{d,w}$. Finally, $q(\eta_d|\gamma_d )$ is Dirichlet with prior parameter $\gamma_d$. 
For topic weights $\eta_d$ and topic labels $z_{d,w}$, we adopt the approach used in the standard LDA model (see \cite{blei2003latent}). In the variational distribution, $\eta_d$ $(d = 1,2,\cdots,D)$ is sampled from a Dirichlet distribution parametrized by the $d-$th row of $\gamma$, a $D\times K$matrix with positive entries. Latent variable $z_{d,w}$ $(w = 1,2,\cdots,n_d)$ is sampled from a multinomial distribution parametrized by a probability vector $\phi_{d,w}$ of length $K$. 

The introduction of keywords entails a new approach on the variational distribution that is amenable to both the categorical nature of $z^{kw}_d$ and the framework of the EM algorithm. Here we use the Gumbel-softmax trick \cite{jang2016categorical} to generate keywords $z^{kw}_d$ in the variational distribution. Specifically, the trainable parameter $\epsilon$ is a $D\times Q$ matrix, where each element $\epsilon_{d,j} \in [0,1]$ represents the probability that the $j-$th candidate keyword is used as a keyword for document $d$. The Gumbel-softmax trick is used to approximate the categorical distribution of $z^{kw}_d$ in the following way where $s_{r}(x)$ is 0 if $x<r$ and 1 otherwise. 

For a fixed temperature parameter $\tau > 0$ and for any given document index $d = 1,2,\cdots, D$, we sample $Q^2$ samples $u_{d,j,k}$ from uniform distribution on $[0,1]$, and we set $q_{d,j,k} = -\log(-\log( u_{d,j,k} ))$ for $j=1,\dots,Q,k=1,
\dots,Q$ to represent a standard Gumbel sample. We then compute
\begin{align}
    y^{kw}_{d,j,k} = \frac{\exp( (\log(\epsilon_{d,j,k}) + g_{d,j,k})/\tau)}{\sum_{i=1}^Q \exp( (\log(\epsilon_{d,i,k}) + g_{d,i,k})/\tau)}  \label{gumbel}
\end{align}
and collect the vector $y^{kw}_{d,k} = (y^{kw}_{d,j,k})_{j = 1}^Q$. We then use $y^{kw}_{d}=s_{0.5}(\sum_{k=1}^Q y^{kw}_{d,k})$ (here $s$ is applied coordinate-wise) to approximate $z^{kw}_d$. Note that if $y^{kw}_{d,k}$ are binary, then $y^{kw}_{d}=s_{1.0}(\sum_{k=1}^Q y^{kw}_{d,k})=s_{0.5}(\sum_{k=1}^Q y^{kw}_{d,k})$. 
The Gumbel-softmax trick provides a differentiable, smooth, empirically efficient and stable gradient estimation approach for categorical distributions, which can recover the original discrete distribution when the temperature is annealed down to $0$. For training of KLDA, we use a sequence of pre-determined, decreasing temperatures. In each iteration we first sample $y^{kw}_d$ from the Gumbel distribution described above, and then feed $y^{kw}_d$ into $f$ and $g$ for evaluating the loss function or computing a gradient during training. 

With the aforementioned variational distribution, we apply the EM algorithm to maximize the Evidence Lower BOund (ELBO) (see \cite{hoffman2010online} for a detailed description based on the original LDA model). In our context, the probability of the corpus $\mathcal{W}$ in KLDA is $p(\mathcal{W}|\theta,\psi,\lambda)$, the joint distribution of the corpus with all latent variables is denoted by $p(\mathcal{W},\mathbf{z}^{kw},\mathbf{\eta},\mathbf{z}|\theta,\psi,\lambda)$, and the probability of latent variables under the variational distributions is represented as $q(\mathbf{z}^{kw},\mathbf{\eta},\mathbf{z})$. Then we have
\begin{align*}
    \log p(\mathcal{W}|\theta,\psi,\lambda) & \geq \mathcal{L}(\theta,\psi,\lambda, \epsilon, \gamma, \phi) \\
    & := \mathbb{E}_q [\log p(\mathcal{W},\mathbf{z}^{kw},\mathbf{\eta},\mathbf{z}|\theta,\psi,\lambda)] \\ & - \mathbb{E}_q[\log  q(\mathbf{z}^{kw},\mathbf{\eta},\mathbf{z}|\epsilon, \gamma, \phi) ], 
\end{align*}
where $\mathbb{E}_q$ denotes the expectation operator with respect to variational distribution $q$. Note that the introduction of the Gumbel-softmax trick affects the variational distribution and requires sampling from the Gumbel distribution at each iteration, and for large-scale cases the training of the model needs to be executed in an online fashion, using only a batch of the entire data in each iteration. Therefore, the actual training is performed based on an estimator $\hat{\mathcal{L}}$ of $\mathcal{L}(\theta,\psi,\lambda, \epsilon, \gamma, \phi)$ depending on the batch sampled in each iteration, and the sample $y^{kw}_d$ through the Gumbel-softmax trick. 

In the remainder, we abuse the notation of $q$ to denote the revised variational distribution where sampling of $z^{kw}$ is performed through the Gumbel trick. First, note that
\begin{align*}\hat{\mathcal{L}} = \mathbb{E}_{\epsilon}\Big[\  \mathbb{E}_q[ & \log p(\mathcal{W},\mathbf{y^{kw}},\mathbf{\eta},\mathbf{z}|\theta,\psi,\lambda)  \\ - & \log  q(\mathbf{y^{kw}},\mathbf{\eta},\mathbf{z}|\ \epsilon, \gamma, \phi)   |\mathbf{y^{kw}}] \Big] 
\end{align*}
where $\mathbb{E}_{\epsilon}$ is the expectation with respect to the random vectors $\mathbf{y^{kw}}$ sampled from the Gumbel distribution parametrized by $\epsilon$. Now we consider each terms one by one, conditioned on $\mathbf{y^{kw}}$. We have
\begin{align}
     & \mathbb{E}_q \Big[\log p(\mathcal{W},\mathbf{y^{kw}},\mathbf{\eta},\mathbf{z}|\theta,\psi,\lambda)|\mathbf{y^{kw}}  \Big]  \nonumber \\
     \approx & \sum_{d = 1}^D  \log p_{kw}(y^{kw}_d | \theta ) \label{termkw}   \\ 
     + & \sum_{d = 1}^D \mathbb{E}_q [ \log p( \eta_d | y^{kw}_d, \psi )] \label{termtopic}  \\
     + & \sum_{d = 1}^D\sum_{w = 1}^{n_d}\mathbf{1}\{ W_{d,w}\notin I(y^{kw}_d) \}\cdot \mathbb{E}_q[ \log p(z_{d,w}|\eta_d) ] \label{termtopiclabel}  \\
     + & \sum_{d = 1}^D\sum_{w = 1}^{n_d}\mathbf{1}\{ W_{d,w}\notin I(y^{kw}_d) \}\cdot\mathbb{E}_q[\log p_w(W_{d,w}|z_{d,w},\beta_d) ] \label{termwords} 
\end{align}
where $I(y) = \{i: y_i > h\}$ for hyperparameter $h$ set to 0.5 in experiments. The indicator functions in \eqref{termtopiclabel} and \eqref{termwords} are capturing the fact that if a word has already been generated as a keyword, we do not need to generate this word again as a regular word; we use hyperparameter $h$ for $y^{kw}_d$ to determine if a word has been generated as a keyword. Note that this is an approximation  only because we are using the Gumbel trick and thus values of $\mathbf{y^{kw}}$ are not strictly $0$ or $1$. 

A direct calculation gives an explicit form for each term \eqref{termkw}-\eqref{termwords}. For term \eqref{termkw}, we have
$$ \log p_{kw}(y|\theta) = \log \tilde{p}_{kw}(y|\theta) - \log Z(\theta)$$
where the unnormalized probability $\tilde{p}_{kw}$ and partition function $Z$ are defined in the previous section. For term \eqref{termtopic}, we have
$$\mathbb{E}_q\log p(\eta_d|y^{kw}_d,\psi) = \tilde{h}(\gamma_d, f_d) $$
where $f_d = f_{\psi}(y^{kw}_d)$ with $\gamma$ being the parameter for the variational distribution, and 
\begin{align*}\tilde{h}(u,v) = & \sum_i (v_i - 1)[\Psi(u_i) - \Psi(\sum_j u_j) ]\\
& + \log \Gamma(\sum_i v_i) - \sum_i \log \Gamma(v_i)\end{align*}
and $\Psi$ denoting the Digamma function. For term \eqref{termtopiclabel} given word with index $w$ in document $d$ we have
$$\mathbb{E}_q[ \log p(z_{d,w}|\eta_d) ] = \sum_{k = 1}^K \phi_{d,w,k}\Big( \Psi(\gamma_{d,k}) - \Psi(\sum_{l = 1}^K \gamma_{d,l} ) \Big)$$ with $\phi$ coming from the variational distribution.
For term \eqref{termwords} we derive
\begin{align*}
   & \mathbb{E}_q[\log p_w(W_{d,w}|z_{d,w},\beta_d) ] \\
    = & \sum_{v = 1}^V\sum_{k = 1}^K \mathbf{1}\{W_{d,w} = v\}\phi_{d,w,k}\Big( \Gamma(\beta_{d,k,v}) - \Gamma(\sum_{j = 1}^V\beta_{d,k,j}) \Big)
\end{align*}
where $\phi$ is the parameter for $z$ in the variational distribution, and $\beta$ is the output of model $g$.

Also, for the variational distribution we have
\begin{align}
    & \mathbb{E}_q[\log  q(\mathbf{y^{kw}},\mathbf{\eta},\mathbf{z}|\ \epsilon, \gamma, \phi)  ] \nonumber \\
    \approx & \sum_{d = 1}^D \log q(y^{kw}_d\ | \ \epsilon_d) \label{termepsilon}  \\
    + & \sum_{d= 1}^D \mathbb{E}_q[ \log q(\eta_d|\gamma_d ) ] \label{termentropytopic}  \\
    + & \sum_{d = 1}^D\sum_{w = 1}^{n_d}\mathbf{1}\{ W_{d,w}\notin I(y^{kw}_d) \}*\mathbb{E}_q[ \log q(z_{d,w}\ | \ \phi_{d,w}) ] \label{termentropyz} .
\end{align}
Specifically, in term \eqref{termepsilon}, given the original Bernoulli distribution in Figure \ref{fig_graph_model} right, we have
$$\log q(y| \epsilon) = \sum_{j = 1}^Q y_j\log(\epsilon_j) + (1 - y_j)\log(1 - \epsilon_j) $$
for vectors $y,\epsilon$ of length $Q$; for term \eqref{termentropytopic} we derive
$$ \mathbb{E}_q[\log q(\eta | \gamma_d)] = \tilde{h}(\gamma_d,\gamma_d);$$
and for term \eqref{termentropyz} we have
$$ \mathbb{E}_q[ \log q(z_{d,w}| \phi_{d,w} ) ] =\sum_{k = 1}^K \phi_{d,w,k}\log\phi_{d,w,k}. $$

We next consider the case where we can only access a batch $\tilde{\mathcal{W}}\subset\mathcal{W}$ of the entire corpus. Denote the number of documents in $\tilde{\mathcal{W}}$ as $B = |\tilde{\mathcal{W}}|$, and the total number of words of the documents in this batch as $N_{B}$. Then we can use the following estimator with adjusted weights for each term: 
\begin{align}
    & \hat{\mathcal{L}} \label{totalloss}\\
    = & \frac{D}{B}\sum_{d \in \tilde{\mathcal{W}}}\log p_{kw}(y_{d}^{kw}|\theta) \nonumber \\
    + & \frac{D}{B}\sum_{d \in \tilde{\mathcal{W}}} \mathbb{E}_q [ \log p( \eta_d | y^{kw}_d, \psi )] \nonumber\\
    + & \frac{V}{N_{B}}\sum_{d \in \tilde{\mathcal{W}}}\sum_{w = 1}^{n_d}\mathbf{1}\{ W_{d,w}\notin I(y^{kw}_d) \} \cdot \mathbb{E}_q[ \log p(z_{d,w}|\eta_d) ] \nonumber\\
    + & \frac{V}{N_{B}}\sum_{d \in \tilde{\mathcal{W}}}\sum_{w = 1}^{n_d}\mathbf{1}\{ W_{d,w}\notin I(y^{kw}_d) \} \nonumber\\ & \qquad\qquad\qquad\qquad \cdot\mathbb{E}_q[\log p_w(W_{d,w}|z_{d,w},\beta_d) ] \nonumber\\
    - & \frac{D}{B}\sum_{d \in \tilde{\mathcal{W}}} \log q(y^{kw}_d\ | \ \epsilon_d) \nonumber\\
    - & \frac{D}{B}\sum_{d \in \tilde{\mathcal{W}}}\mathbb{E}_q[ \log q(\eta_d|\gamma_d ) ] \nonumber\\
    - & \frac{V}{N_{B}}\sum_{d \in \tilde{\mathcal{W}}}\sum_{w = 1}^{n_d}\mathbf{1}\{ W_{d,w}\notin I(y^{kw}_d) \}\cdot\mathbb{E}_q[ \log q(z_{d,w}\ | \ \phi_{d,w}) ]. \nonumber
\end{align}

At an E-step, the optimal solution with respect to $\gamma$ and $\phi$ can be solved through the iterative updates for any $d \in \tilde{\mathcal{W}}$ and $k = 1,2,\cdots,K$ as
\begin{align*}
    \phi_{d,w,k} & \propto \beta_{d,k,W_{d,w}}\exp( \Psi(\gamma_{d,k}) - \Psi(\sum_{j = 1}^K\gamma_{d,j}) ) \\ & \qquad \qquad \qquad \text{for all } W_{d,w}\notin I(y^{kw}_d); \\
    \gamma_{d,k} & = \big( f_{\psi}( y^{kw}_d ) \big)_k + \sum_{w = 1}^{n_d}\phi_{d,w,k}.
\end{align*}
The derivation can be found in \cite{blei2003latent}. For all the other parameters, the updates can be conducted through gradient descent with respect to the estimated log probability $\hat{\mathcal{L}}$. Note that $\epsilon$ can also be updated in this fashion since $y^{kw}_d$ is essentially a function of $\epsilon$ (and random uniform samples $u$).

\subsection{Partition Function}\label{sec:partitionfunction}

Updating of $\theta$ requires special care with respect to $\log Z(\theta)$ in \eqref{termkw}. The exponential nature of this term also requires approximations in inference. We resort to sampling described next. 

For estimating the gradient $\nabla_{\theta}\log Z(\theta)$ we use the following well-known fact (see \cite{Goodfellow-et-al-2016} for details)
$$ \nabla_{\theta}\log Z(\theta) = \mathbb{E}_{X \sim p_{kw}}  \nabla_{\theta} \log \tilde{p}_{kw}(X).$$
During training, the gradient estimator is obtained through the Markov Chain Monte Carlo algorithm. Specifically, we employ Gibbs sampling, and maintain $N_{MCMC}$ different independent chains whose state spaces are $\{0,1\}^Q\setminus \{0\}$, the possible range of $z^{kw}$. Before training starts, we initialize all the chains by uniformly randomly sampling a possible $z^{kw}$. At iteration $t$ of gradient optimization in the M-step, we perform $m_t$ steps of Gibbs sampling for each chain, and denote the current state for each chain by $\tilde{z}_{t}^{(j)}$ for each $j = 1,2,\cdots,N_{MCMC}$. Here $\{m_t\}_{t \geq 1}$ is a preset sequence of integers. The gradient estimator we use to perform stochastic gradient descent on $\theta$, conditioned on $y^{kw}$ from the Gumbel trick, is
\begin{align}
    \frac{D}{B}\sum_{d \in \tilde{\mathcal{W}}}\Big(\nabla_{\theta}\log \tilde{p}_{kw}(y_{d}^{kw}|\theta) - \frac{\sum_{j = 1}^{N_{MCMC}}\log \tilde{p}_{kw}(\tilde{z}^{(j)}_{t}|\theta)}{N_{MCMC}} \Big). \label{Gestimator} 
\end{align}

During inference with respect to the $\log Z(\theta)$ term, we employ a simple importance sampling algorithm.
Alternative options of MCMC or acceptance-rejection are not appropriate for the following reasons. 
Since the estimate $\hat{Z}$ is used in the log function as an approximation to $\log(Z)$, even though $\mathbb{E}\hat{Z} = Z$, it holds that $\mathbb{E}\log(\hat{Z})\neq \log(\mathbb{E}\hat{Z})$ due to the Jensen's inequality, and the gap is affected by the variance of $\hat{Z}$ and thus we need a lot of samples (we use 100,000). For MCMC running that many chains till burn-in is too costly. 

To be specific, importance sampling is used to estimate the partition function $Z$, where we construct the following estimator
$$\hat{Z} = \frac{1}{N_{e}}\sum_{j = 1}^{N_e}\frac{\tilde{p}_{kw}(X^{(j)})}{h(X^{(j)})}.$$
Values $\{X^{(j)}\}$ are i.i.d samples from distribution $h(\cdot)$ constructed in the following way.
\begin{enumerate}
    \item Sample $\tilde{N}$ from Geometric distribution with mean 1/0.25, and set $N = \min\{\tilde{N},Q\}$ to correspond to the number of keywords;
    \item Randomly sample a subset $\tilde{Q} \subset \mathcal{Q}$ of size $N$ (the sampling is without replacement); 
    \item Output a vector $X$ of length $Q$, where $X_l = 1$ for $l \in \tilde{Q}$ and $X_l = 0$ otherwise.
\end{enumerate}

If the distribution of $h$ is relatively close to the distribution of $Z$, the variance of the estimator should be low. 
The designed $h$ is trying to match the generative process. Even though we are ignoring the actual dependency on $\theta$ for each candidate keyword and use uniform distribution instead, the geometric distribution is capturing the fact that, due to the penalty term $c>0$ in the definition of $\tilde{p}_kw$, it is approximately geometrically or exponentially unlikely to have many keywords.

Note that $h(X)$ is simple to evaluate since it only relies on the number of positive elements in $X$, and only requires the summation of a geometric sequence. In our experiments, we set $N_e = 100,000$ so that the variance of $\log (\hat{Z})$ is small enough when plugging in the estimator $\hat{Z}$, which further ensures that the bias to $\log Z(\theta)$ is small enough. 

\subsection{Overall Training Algorithm}

The overall algorithm is presented in Algorithm \ref{mainalgo}. The main loop in step 2 iterates over all EM iterations. In the inner loop 3 we first create independent samples in support of keyword candidates based on Gibbs Sampling described in Section \ref{sec:partitionfunction}. The actual samples are derived in step 6 for the select batch samples of documents. The forward step is executed in step 7. Possible several E steps are done in step 8 while a single M-step gradient adjustment is performed in steps 11-15. 

\begin{algorithm}[ht]
  \caption{EM algorithm for KLDA model}
  \begin{algorithmic}[1]
    \Require
      \Statex Corpus $\mathcal{W}= \{d_1,d_2,\cdots,d_D\}$; Vocabulary $\mathcal{V}$; Candidate keyword set $\mathcal{Q}\subset\mathcal{V}$; 
      \Statex Learning rates $\{\alpha_t\}_{t\geq 1}$; Temperature $\{\tau_t\}_{t \geq 1}$; MCMC repetitions $\{m_t\}_{t\geq 1}$; Integer for number of MCMC chains $N_{MCMC}$; Batch size $B$; E-step iterations $K_{E}$
    \Statex
\State \textbf{Initialize: } Parameters $\psi,\lambda$ for model $f$ and $g$; Parameters $\theta,\epsilon,\gamma,\phi$ for the probabilistic model; States for MCMC chains $\{\tilde{z}^{(j)}_0\}_{j = 1}^{N_{MCMC}}$

\For{$t = 1, 2, 3, \ldots $}

\For{$ j = 1,2,\ldots,N_{MCMC}$} \Comment{MCMC} 
\State Perform $m_t$ Gibbs Sampling steps on $\tilde{z}^{(j)}_{t-1}$ to \WRP obtain  $\tilde{z}^{(j)}_{t}$
\EndFor

\State Uniformly sample a batch $\tilde{\mathcal{W}}\subset\mathcal{W}$ of size $B$

\State Obtain $y^{kw}_d$ for any $d \in \tilde{\mathcal{W}}$ via \eqref{gumbel} with \WRP temperature $\tau_{t}$ \Comment{Gumbel Trick}
\State Evaluate $\hat{\mathcal{L}}$ using \eqref{totalloss}

\For{$\tilde{t} = 1,2,\ldots,K_E$} \Comment{E-step}
\State $\phi_{d,w,k}  \propto  \beta_{d,k,W_{d,w}}\exp( \Psi(\gamma_{d,k}) - \Psi(\sum_{j = 1}^K\gamma_{d,j}) )$ \WRP $ \text{for each } W_{d,w}\notin I(y^{kw}_d) $ \WRP $\text{and for each } d \in \tilde{\mathcal{W}};$
 \State  $\gamma_{d,k}  = \big( f_{\psi}( y^{kw}_d ) \big)_k + \sum_{w = 1}^{n_d}\phi_{d,w,k}\ \forall d \in \tilde{\mathcal{W}};$
\EndFor

\State $\epsilon \leftarrow \epsilon + \alpha_t \nabla_{\epsilon}\hat{\mathcal{L}}$ \Comment{M-step}

\State $\psi \leftarrow \psi + \alpha_t \nabla_{\psi}\hat{\mathcal{L}}$ 

\State $\lambda \leftarrow \lambda + \alpha_t \nabla_{\lambda}\hat{\mathcal{L}}$

\State Evaluate $\nabla_{\theta}\hat{\mathcal{L}}$ using \eqref{Gestimator}

\State $\theta \leftarrow \theta + \alpha_t \nabla_{\theta}\hat{\mathcal{L}}$
\EndFor
    
  \end{algorithmic} \label{mainalgo}
\end{algorithm}

\section{Inference}\label{secInference}

After training the KLDA model, we obtain model $g_{\lambda}$ that can output the topic-word matrices given keyword set $z$, and vector $\theta$ that can be used to evaluate the possibility that keyword set $z$ is actually generated by the KLDA procedure. The next goal is to decide the best possible keyword sets for further collection of documents. To this end, we propose two inference metrics that aim to capture two antagonistic purposes of deriving a keyword set. Let us recall that the documents are assumed to be collected in time (which is actually disregarded in training). Let $\mathcal{Q}_{last}$ be the most recent set of keywords used to collect documents $\mathcal{D}_{last}=\{d_{D-s},\cdots, d_D\}$ for a hyperparameter $s$. In practice $s$ should reflect the frequency of keyword updates, e.g., if we  plan to update the keywords once every week, then these documents should correspond to the documents in the last week and thus $\mathcal{Q}_{last}$ are the keywords used in the previous week. 

Given a subset $U\subset \mathcal{Q}$ of keywords, then $g_{\lambda}(1_{U})$ is a topic-word matrix where 
$1_U\in \{0,1\}^Q$ is the binary representation of $U$. Let $a(U)=\sum_{k} g_{\lambda}(1_{U})_k$ / K is a distribution over vocabulary (with $g_{\lambda}(1_{U})_k$ denoting the $k$-th row of matrix $g_{\lambda}(1_{U}))$. Let also $\text{freq}(S)$ be the ratio of the number of documents in $\mathcal{Q}_{last}$ containing every word in $S$ and $s$ (the number of all documents in $\mathcal{Q}_{last}$).

To be more specific about the goal, assuming that a set $\mathcal{Q}_{last} \subset \mathcal{Q}$ is provided, the goal is to identify another subset $\mathcal{Q}_{next} \subseteq \mathcal{Q}$ that, in terms of the possible documents generated by keywords contained in $\mathcal{Q}_{next}$, it maintains information in documents $\mathcal{D}_{last}$ corresponding to $\mathcal{Q}_{last}$ but also captures new information and provides a natural extension of previous topics.

We construct $\mathcal{Q}_{next}$ by considering each word $W_{last}\in \mathcal{Q}_{last}$ and try to ``extend'' it with a different word $W_{next}\in \mathcal{Q}$. The set of all derived $W_{next}$ form our final set $\mathcal{Q}_{next}$ of keywords. Let $W_{last},W_{next}$ be fixed in the subsequent discussion. We consider the strength of $W_{next}$ by using the following two metrics. 



The first metric aims to capture that $\{W_{next},W_{last}\}$ can be seen as a good extension of keyword $W_{last}$, since either of them generates documents that cover similar topics. We need to capture that as long as one of the keywords appears in the document it would be meaningful to collect it. Choices for such extensions $W_{next}$ should satisfy the following criteria: (1) $W_{next}$ should be relatively frequent in $\mathcal{D}_{last}$ (we set the threshold as $f_2=0.5\%$), otherwise the mixing of $\{W_{next},W_{last}\}$ would be extremely similar to using $W_{last}$ only, thus rendering the next criterion meaningless; (2) the KL divergence between the word distributions induced by $W_{last}$ and $W_{next}$ should be small ($KL(a(\{W_{last}\})|a(\{W_{next}\}))$ should be small where KL denotes KL divergence). In our experiments we found that, instead of using KL divergence, most other $f-$divergence measures can be used and the selected keywords remain largely similar, so other divergence metrics can be used if need be. As long as the first threshold is satisfied, we prefer the keywords that induce as small KL divergence as possible.

Besides this metric, we propose the second metric that is used as a reference and helps filtering out some obviously bad choices before applying the previous metric. We use the term \textit{high-frequency distance} for the quantity defined as follows. Let us assume we have hyperparameters $o_h \in [0,1]$ (high-frequency threshold), $r \in [0,1]$ (retention rate), and $\hat{c} > 0$ (penalty weight). 
We define the following sets
\begin{itemize}
    \item extension set $$E = \{w \in \mathcal{V}: a(\{W_{next}\})_w > o_h, \frac{a(\{W_{last}\})_w}{a(\{W_{next}\})_w}<r\}$$
    \item retention failure set $$F = \{w \in \mathcal{V}: a(\{W_{last}\})_w > o_h, \frac{a(\{W_{next}\})_w}{a(\{W_{last}\})_w}<r \}$$
\end{itemize}
and set the high-frequency distance as $R(W_{last},W_{next})=|E| - \hat{c}|F|$. Intuitively, this quantity translates the idea of ``maintaining old information but also extending to new information'' into a value that is based on high-frequency words in both cases. In our experiments we set $o_h$ to be a value that ensures that the high-frequency sets $$\{w: a(\{W_{next}\})_w > o_h\}, \{w: a(\{W_{last}\})_w > o_h\}$$ are of size around 100 (the value $o_h$ is set as $1/1,000$), and we set $\hat{c} = 1, r = 1/2$. Note that if $W_{last}=W_{next}$, then the two sets are empty and thus 
$|E| - \hat{c}|F|=0$. 

The actual inference procedure is presented in Algorithm  \ref{algoinference}.
Note that $S\subset C$. It can conceptually happen that $\mathcal{Q}_{next}=\emptyset$ is returned by the algorithm. In such a case which was never observed in the experimental study, we can select
$\mathcal{Q}_{next}=\mathcal{Q}_{last}$.

\begin{algorithm}[ht]
  \caption{Inference}
  \begin{algorithmic}[1]
  \State $\mathcal{Q}_{next} = \emptyset$
  \For{$W_{last}\in \mathcal{Q}_{last}$}
    \State $C=\{W_{next} \in \mathcal{Q} |                   R(W_{last},W_{next}) \ge 0\}$
    \State $S = \text{argmin}_{W_{next} \in C: \text{freq}(\{W_{next}\})>f_2}$ \WRP $\{
    KL(a(\{W_{last}\})|a(\{W_{next}\}))
    \}$
    \State $\mathcal{Q}_{next} = \mathcal{Q}_{next} \cup S$
  \EndFor
  \State Remove duplicate words in $\mathcal{Q}_{next}$
  \end{algorithmic}\label{algoinference}
\end{algorithm}
  
\section{Computational Study}

\subsection{Training Details}

Related to (\ref{gumbel}), function $s$ is not smooth and it requires $Q^2$ samples. In order to circumvent this in the computational experiments we instead sample $Q$ binary Gumble approximations as follows.
For any given document index $d = 1,2,\cdots, D$, and keyword index $j = 1,2,\cdots,Q$, we sample two independent copies $u_{d,j,1},u_{d,j,0}$ from the uniform distribution on $[0,1]$,  and we set $g_{d,j,i} = -\log(-\log( u_{d,j,i} ))$ for $i = 0,1$. We then compute
\begin{align}
    y^{kw}_{d,j} = \frac{\exp( (\log(\epsilon_{d,j,1}) + g_{d,j,1})/\tau)}{\sum_{i \in \{0,1\}}\exp( (\log(\epsilon_{d,j,i}) + g_{d,j,i})/\tau)}
\end{align}
and define the vector $y^{kw}_{d} = (y^{kw}_{d,j})_{j = 1}^Q$ to approximate $z^{kw}_d$. While this is smooth it has the drawback that it allows $y^{kw}_{d}=0$, i.e., no keyword is in document $d$. Thus this necessitates allowing state 0 in MCMC (see Section \ref{sec:partitionfunction}).

For our experiments, we set $K = 5$ topics (by finding a good number of topics using plain LDA), and the batch size $B = 64$. For the distribution of $p_{kw}$, we set the penalty $c = 2$.  For model $g$ (topic-word matrices) we use a 2-layer neural network with 32 neurons inn each layer, using LeakyRelu activation with weight $0.01$ for the negative quadrant (see \cite{maas2013rectifier} for details on LeakyRelu). For model $f$, we fix it and always use a uniform $\mathbf{1}$ vector as the output. In other words, in our experiments we use an uninformative, uniform prior for topic weights. 
We tried to use a parametric $f$ but we observed that 
the trained model tends to use only one of the $K$ topics and the others have an extremely small probability to be generated. We have tried to add a Lagrangian penalty term to prevent the topic weights from being too extreme but at no avail. For training of model $g$ and gradient ascent on most parameters, we set the learning rates as $0.001$ and use the Adam optimizer \cite{kingma2014adam}. For updates of $\epsilon$, however, we use RMSprop (see \cite{zou2019sufficient}) with learning rate $0.005$ to cope with the fact that the gradient on $\epsilon$ is extremely sparse as only the $\epsilon$ corresponding to a mini batch of documents is updated at each iteration. We also add an L2 regularization term to loss function (\ref{totalloss}) with the regularization penalty of 0.1.

For the MCMC steps, we set $m_t = 1000$ if $t\mod{100} = 0$ and set $m_t = 1$ otherwise. In other words, periodically we perform more Gibbs sampling steps to ensure the quality of the samples after updating $\theta$. For the Gumbel trick, we anneal down the temperature based on the following scheme
$$\tau_t = \max\{0.25, \exp( -0.0001*t ) \}, $$
which is partially following the choices in \cite{jang2016categorical}
but we set the threshold as $0.25$ as we try to further anneal down the temperature and better recover the original categorical distribution.

Since the corpus is given, during training of KLDA, we cannot allow the model to generate a keyword that is not observed in the document. To this end, at each iteration after we obtain $y^{kw}_d$ through the Gumbel trick, we apply a binary mask to all the elements in $y^{kw}_d$, which multiplies $0$ to slots of candidates keywords that do not appear in the document. Therefore, the masked candidate keywords do not affect the evaluation of $\hat{\mathcal{L}}$, and the corresponding $\epsilon$ does not need to be updated. 

Training of $\theta$ is critical to our inference, since it determines the process that generates keywords. In our experiment we apply several tricks to improve the quality of $\theta$. First, we change the weights of the negative MCMC samples for estimating the gradient of the partition function. In \eqref{Gestimator} we apply a weight $c_{\text{neg}}$ and use the revised form
$$ \nabla_{\theta} \log \tilde{p}_{kw}(y^{kw}_d|\theta) - c_{\text{neg}} \frac{\sum_{j = 1}^{N_{MCMC}}\log \tilde{p}_{kw}(z^{(j)}_{t}|\theta)}{N_{MCMC}}.$$
In our experiment we set $c_{\text{neg}} = 0.1$. Second, we add an $L_2$ regularization term in $\hat{\mathcal{L}}$ as the squared error between normalized $\exp(\theta)$ and the empirical probability of each candidate keyword in the corpus. We set the coefficient of this $L_2$ regularization term as 1.

Lastly, to improve the quality of model $g$, we adopt a pre-training scheme before  training the KLDA model. The only difference between pre-training and complete KLDA model training is in the following. In pre-training, there is no generation of keywords $z^{kw}$; instead we use the real observation as $z^{kw}$, meaning that each element in $z^{kw}$ is $1$ if the corresponding keyword appears in the document, and $0$ otherwise (namely, we skip the stochastic generation of keywords, and deterministically generate a candidate keyword as a keyword if it is present in the document). We use the same hyperparameters and model structures as described above. Note that this pre-training step does not involve updates on $\theta$ (since in fact we skip the keyword generation step), and only serves to improve the quality of topic-word matrices output by $g$. We run pretrain for $2,500$ iterations, and when we start training the full KLDA model, we load the parameters of model $g$ from pretrain to initialize the new $g$ model.

\subsection{Data}

\begin{table*}[t]
  \centering
  \caption{Evolution of Keywords, Weeks 1-10}
\label{table:KeywordsEvolution}
  \begin{tabular}{c|c|c|c}
  \hline
  Week & \textbf{Horse-racing} & \textbf{Streaming} & \textbf{KPOP} \\
  \hline
  1 & computer virus, Trojan, invasion, malware & video, stream & idol, album \\
  \hline 
  2 & horse, sex, virus, aid, war & league, hero, game, live, playing & ticket, army, follow \\
  \hline
  3 & horse, sex, virus, war & stream, hero, game, live, playing & ticket, army \\
  \hline 
  4 & horse & stream, game, live, playing & really, army \\
  \hline
  5 & horse & stream, game, live, playing & army, known, stan \\
  \hline
  6 & horse, happy & stream, game, live, playing & persona, april \\
  \hline 
  7 & people, know & stream, live, playing & love, april\\
  \hline
  8 & people, racing & stream, live, playing, time & album, april, army, luv \\ 
  \hline 
  9 & people, racing & stream, live, playing, workshop & army \\
  \hline
  10 & derby, race, racing & stream, game, playing & army \\
  \hline
  \end{tabular}
\end{table*}

\begin{table*}[t]
  \centering
  \caption{Additional Keywords for Week 11}
\label{table:AdditionalKWWeek11}
  \begin{tabular}{c| p{25em} |p{17em}}
  \hline
   & Recommendations by \textit{viral} & High-frequency Words \\
  \hline
  \textbf{Horse-racing} & birthday Kentucky derby shit racing riding people today time today & maximum security town one girl show \\
  \hline 
  \textbf{Streaming} & stream skin map mercy game havana league player night tonight people fun & video guy everyone \\
  \hline 
  \textbf{KPOP} & army idol concert persona stan world day tour people time way & thank everyone album guys \\
  \hline
  \end{tabular}
\end{table*}

We next give a briefing of the data used in the experiments. In order to test the model's capability under different settings, we use collections of Twitter posts that concern various topics and evolve over time. Specifically, data collection is done through the Twitter streaming API; the entire collection lasted for 11 weeks in year 2019 for each dataset with keywords adapted at the beginning of each week. During this period we maintain 3 sets of tweets revolving around 3 different topics: horse-racing, streaming, and Korean pop music (the first dataset was intended to be a ``computer virus" related dataset, but evolved into a horse-racing dataset through the keyword selection benchmark procedure described next). We selected these topics to have variety. 

The benchmark algorithm relies on a prediction model that classifies if a tweet is likely to become viral or not. The features of a tweet are the user identification, text length, number of URLs, number of mentions, number of hashtags and emojis, the number of users the user is following, the number of followers of the user, the number of favorites of the user, and the total number of posts of the user. The label is defined as viral for each tweet that has been retweeted at least ten times in a month and the model is trained based on 3 months of tweets. 

In terms of the data collection procedure for any dataset, in the first week we manually pick a set of keywords that are related to these topics and, by querying the API, collect tweets posted in the following 7-day period that contain any of the given keywords. 
For all subsequent weeks we apply the following algorithm called \textit{viral} to adjust the set of keywords. The algorithm also serves as the main benchmark algorithm.  
\begin{itemize}
\item We first calibrate an LDA model on all tweets from the last week. 
\item We next perform inference of the tweets collected in the last week by using the viral prediction model. 
\item From all tweets predicted to become viral, we select 15 highest count words (excluding words that are not nouns or proper nouns which implies that we first perform part-of-speech tagging) as keyword candidates. 
\item Let $tweets(k)$ be the set of all tweets in the previous week containing keyword $k$ where $k$ is one of the candidate keywords based on the viral tweets. 
\item For each keyword $k$ of the 15 keywords and for each tweet in $tweets(k)$ we check if the tweet belongs to one of the topics. This is done by using the LDA probabilities. If over 60\% of the tweets in $tweets(k)$ belong to a topic, keyword $k$ is selected and otherwise it is discarded. 
\end{itemize}
The `viral' component produces candidate keywords present in tweets that are likely to become important. The LDA part is necessary to assure that the selected keywords are not deviating too much from the desired topics. For example, in the case with the industry partner, the company had a promotion related to the FIFA World Cup. If the LDA part is removed, then the viral tweets capture everything related to the World Cup which is too broad and most of the material of no interest to the company. 

We execute the algorithm for each of the remaining 9 weeks for each one of the 3 datasets. The keywords produced and thus used in the collection process are exhibited in Table \ref{table:KeywordsEvolution}. We remark that based on this algorithm the initial topic of ``computer virus'' evolved to horse-racing in the first dataset. 

Next, based on the tweets collected in this 10-week span, we generate a set of words that are the union of the following: (1) recommendation from \textit{viral} using the tweets in week 10; (2) high-frequency nouns in tweets not recommended by \textit{viral}; (3) any candidate keyword that has been used in the 10-week data collection period. This candidate keyword set is used to collect tweets for one more week, labeled as week 11, and is utilized as the test set in our experiment. Table \ref{table:AdditionalKWWeek11} lists the keywords for week 11 that are added to the keywords in Table \ref{table:KeywordsEvolution} and Table \ref{table:DatasetInfo} provides basic statistics of the datasets. KLDA for week 11 only recommends a subset of the keywords shown in Tables \ref{table:KeywordsEvolution} and \ref{table:AdditionalKWWeek11}. 
At the end of week 10, we assemble the keywords for week 11 and then collect all tweets during week 11 based on all these keywords. 

\begin{table*}[!ht]
\centering
\caption{Information of Datasets Used in Experiments}
\label{table:DatasetInfo}
 \begin{tabular}{c|r|r|r} 
\hline
  & \textbf{Horse-racing} & \textbf{Streaming} & \textbf{KPOP} \\ 
 \hline
 \#Docs, Weeks 1-10 & 153,378 & 135,506 & 163,013 \\ 
 \hline
 \#Words, Weeks 1-10 & 1,610,218 & 1,387,365 & 1,817,409 \\
 \hline
 \#Docs, Week 11 & 13,433 & 6,271 & 12,327 \\
 \hline
 \#Words, Week 11 & 136,973 & 64,002 & 12,7454 \\
 \hline
 Vocabulary & 172,371 & 196,299 & 162,639 \\
 \hline
 Vocabulary(reduced) & 10,164 & 12,033 & 12,422 \\
\hline
\end{tabular}
\end{table*}

We add a few remarks regarding collection and preprocessing of the datasets. First, we apply stemming and removing of the stopwords. Besides, we reduce the vocabulary size by replacing all the low-frequency tokens (appearing less than 10 times in all tweets collected during the first 10 week) by a placeholder token. The purpose is to facilitate model training by preventing rare typos and acronyms, since tweets are more prone to these issues. As we observe in the collected datasets, less than 10\% of tokens (words) have appeared 10 or more times in the entire dataset. By substituting the low-frequency tokens with the placeholder token, we manage to reduce the vocabulary to a reasonable size (see Table \ref{table:DatasetInfo} for details). Lastly, if a token outside of the vocabulary is observed in the tweets collected in week 11, it is also replaced by the placeholder token, thus maintaining a consistent vocabulary between weeks 1-10 and week 11.

\subsection{Computational Results}

\begin{table*}[t]
\centering
\caption{Perplexity of Trained Models}
\label{table:perplexity}
 \begin{tabular}{c|c|c|c} 
\hline
  & Horse-racing & Streaming & KPOP \\ 
 \hline
 LDA on Week 10 & -4.86 & -4.94 & -4.42 \\
 \hline
 Trained on Weeks 1-9, Tested on Week 10 & -6.46 & -5.92 & -6.889  \\
 \hline
 no training at all & -8.30 & -7.47 & -8.51 \\
 \hline
\end{tabular}
\end{table*}

\begin{table*}[t]
\centering
\caption{Recommendation of Keywords from KLDA Model: Individual Keyword Level}
\label{table:recommendations_kwLevel}
 \begin{tabular}{c|c|c|c|c|c|c} 
\hline
 Org. Keyword & KLDA (Top 2) & KLDA (Top 3) & Week 11 (Top 2) & Week 11 (Top 3) & Acc. Top 2 & Acc. Top 3\\ 
 \hline\hline
derby & racing Kentucky & racing Kentucky time & racing sex & racing sex know & 0.50 & 0.33 \\ 
 \hline
 race & war riding & war riding horse & horse riding & horse riding racing & 0.50 & 0.67 \\
 \hline
racing & today kentucky & today kentucky time & sex know & sex know show & 0.00 & 0.00 \\
 \hline \hline
 stream & map & map & none & none & &\\
 \hline
 game & playing tonight & playing tonight stream & playing tonight & playing tonight fun & 1.00 & 0.67 \\
 \hline
 playing & live stream & live stream map & fun tonight & fun tonight stream & 0.00 & 0.33 \\
 \hline \hline
 army & april album & april album & fans concert & fans concert guys & 0.00 & 0.00 \\
 \hline
\end{tabular}
\end{table*}

We outline the objective of the computational experiments as follows: for each dataset, different methods yield different sets of recommendations of keywords for week 11; by collecting tweets using candidate keywords recommended by different methods at week 11, we are able to evaluate the performance of different methods based on tweets in week 11 (given a set of keywords for week 11, we simply select the tweets containing at least one keyword from the set; since such keywords are a subset of all keywords used in the collection process for week 11, no tweet is left out); the comparison of different methods under various topics also demonstrate the versatility and stability of different approaches.

More specifically, we compare KLDA against \textit{viral} and a randomly chosen keyword set. Recall that the inference algorithm based on KLDA generates recommendations for a given keyword by evaluating all the remaining candidate keywords under the aforementioned metric. Based on the score of the metric, we need to specify a threshold: candidate keywords with metric scores meeting the threshold shall be accepted into the recommendation list. In this experiment we compare the results from two different thresholds: we either accept the top 2 words or the top 3 words. Throughout, we use \textit{top 2} or \textit{top 3} to refer to the two different set of criteria, respectively. As for \textit{viral}, recall that the recommendation is generated for the entire set used in week 10 in one shot, instead of extending each keyword in the set as is the case in Algorithm \ref{algoinference}. Lastly, to generate a \textit{random} keyword set as another benchmark, we randomly choose 3 keywords used in the week 11 candidate set, with the restriction that only those that are not recommended by Algorithm \ref{algoinference} or \textit{viral} are taken into consideration. The random set generated for the experiment is: Horse-racing: town, trojan, know; Streaming: workshop, everyone, hero; KPOP: know, everyone, luv.

\begin{table}[h!]
\centering
\caption{Performance of Different Recommendations}
\label{table:performance}
 \begin{tabular}{c|c|c|c} 
\hline
 Top 2 & KLDA & viral & Random \\
 \hline
 Horse-racing, Accuracy & \textbf{0.40} & 0.20 & 0.33 \\
 \hline 
 Horse-racing, Coverage & \textbf{0.40} & \textbf{0.40} & 0.20 \\
 \hline
 Streaming, Accuracy & \textbf{0.50} & 0.25 & 0.00 \\
 \hline
 Streaming, Coverage & 0.67 & \textbf{1.00} & 0.00 \\
 \hline
 KPOP, Accuracy & 0.00 & \textbf{0.09} & 0.00 \\
 \hline 
 KPOP, Coverage & 0.00 & \textbf{0.50} & 0.00 \\
 \hline \hline 
 Accuracy (Mean) & \textbf{0.30} & 0.17 & 0.11 \\
 \hline
 Coverage (Mean) & 0.36 &  \textbf{0.63} & 0.07 \\
 \hline\hline \hline
 Top 3 & KLDA & viral & Random \\
 \hline
 Horse-racing, Accuracy & \textbf{0.43} & 0.20 & 0.33 \\
 \hline 
 Horse-racing, Coverage & \textbf{0.50} & 0.33 & 0.17 \\
 \hline
 Streaming, Accuracy & \textbf{0.60} & 0.33 & 0.00 \\
 \hline
 Streaming, Coverage & 0.75 & \textbf{1.00} & 0.00 \\
 \hline
 KPOP, Accuracy & 0.00 & \textbf{0.09} & 0.00 \\
 \hline 
 KPOP, Coverage & 0.00 & \textbf{0.33} & 0.00 \\
 \hline \hline 
 Accuracy (Mean) & \textbf{0.34} & 0.21 & 0.11 \\
 \hline 
 Coverage (Mean) & 0.41 & \textbf{0.55} & 0.06 \\
 \hline
\end{tabular}
\end{table}

We first demonstrate the predictability of the trained model in terms of topics of future tweets, thus showing KLDA's capability for this modelling problem. Specifically, we first train a model using tweets collected from weeks 1 to 9; for tweets in week 10 we train a separate LDA model; we then compare the perplexity of the two models when fitting the tweets in week 10. As shown in Table \ref{table:perplexity}, we see a consistent pattern across all the three datasets: by training a model using tweets from week 1-9, we are able to fit tweets in the next week with very good perplexity, which is both a significant improvement compared to ``no training at all'' (a model that is randomly initialized and not trained at all), and close to the performance of the LDA model trained on week 10 directly.

As mentioned before, the KLDA inference Algorithm \ref{algoinference} generates recommendations by inspecting each keyword used in week 10. We next discuss the performance of KLDA at the individual keyword level, the result of which is illustrated in Table \ref{table:recommendations_kwLevel}. The metrics accuracy and coverage are interpreted as follows. Calling the top 2 (or 3) recommendations for week 11 based on the weeks 1-10 model as predictions, and the top 2 (or 3) choices in week 11 based on the two criteria discussed in Section \ref{secInference} with respect to LDA based on tweets in week 11 as truth, we claim that a prediction generated by KLDA is accurate if it is in truth, and we say one truth is covered by KLDA if it is in the prediction set. We use ``none'' to indicate that due to bad performance of the high-frequency distance defined in Section \ref{secInference}, all candidate keywords are excluded and there is no good recommendation based on the tweets in week 11. We observe that KLDA demonstrates great accuracy when accepting top 2 choices, and the result further improves when we are allowed to accept top 3 choices.

The comparison of the recommendations for week 11 under the different methods is summarized in Table \ref{table:performance}. We observe that under both top 2 and top 3, KLDA yields the highest average accuracy; besides, KLDA also performs the best for 2 of the 3 topics. In the KPOP dataset, which is the only case KLDA is outperformed by \textit{viral}, note that it is expected that \textit{viral} outperforms since it generates a recommendation set of a much larger size, and even so it produces only one accurate prediction out of 11. In general, it is expected \textit{viral} to have higher coverage since it selects many more keywords. 

\bibliographystyle{plain}
\bibliography{citation.bib}

\end{document}